\documentclass[runningheads]{llncs}
\usepackage{graphicx}
\usepackage{comment}
\usepackage{amsmath,amssymb} % define this before the line numbering.
\usepackage{color}
\usepackage[table,xcdraw]{xcolor}

\usepackage{algorithm}
\usepackage{algorithmic}
\usepackage{capt-of}
\usepackage{booktabs}
\usepackage{array}
\usepackage{pifont}

\usepackage{comment}
\usepackage{gensymb}
\usepackage{tabularx}
\usepackage{multirow}
\usepackage{booktabs}
\usepackage{multirow}
\usepackage{adjustbox}
\usepackage{textcomp}
\newcommand{\etal}{\textit{et al}.}

\newcolumntype{R}[2]{%
    >{\adjustbox{angle=#1,lap=\width-(#2)}\bgroup}%
    l%
    <{\egroup}%
}
\usepackage{array}
\newcommand{\PreserveBackslash}[1]{\let\temp=\\#1\let\\=\temp}
\newcolumntype{C}[1]{>{\PreserveBackslash\centering}p{#1}}

\newcommand{\mysubsubsection}[1]{\vspace{0.15cm} \noindent {\bf #1}:}

\begin{document}
% \linenumbers
\pagestyle{headings}
\mainmatter
\def\ECCVSubNumber{677}  % Insert your submission number here

%\title{Space-GAN: Relational Generative Adversarial Networks for Graph-constrained Indoor Space Planning} % Replace
\title{House-GAN: Relational Generative Adversarial Networks for Graph-constrained House Layout Generation} % Replace with your title

% % INITIAL SUBMISSION 
% %\begin{comment}
% \titlerunning{ECCV-20 submission ID \ECCVSubNumber} 
% \authorrunning{ECCV-20 submission ID \ECCVSubNumber} 
% \author{Anonymous ECCV submission}
% \institute{Paper ID \ECCVSubNumber}
% %\end{comment}
% %******************

% \begin{comment}
\titlerunning{House-GAN: Graph-constrained House Layout Generation}
\author{Nelson Nauata\inst{1} \and
Kai-Hung Chang\inst{2}\and
Chin-Yi Cheng\inst{2}\and
Greg Mori\inst{1}\and
Yasutaka Furukawa\inst{1}}
\authorrunning{N. Nauata et al.}
\institute{Simon Fraser University\\
\email{\{nnauata, furukawa\}@sfu.ca, mori@cs.sfu.ca}\\ \and
Autodesk Research\\
\email{\{kai-hung.chang, chin-yi.cheng\}@autodesk.com}}
% \end{comment}
%******************
\maketitle

\begin{figure}
    \centering
    \includegraphics[width=\textwidth]{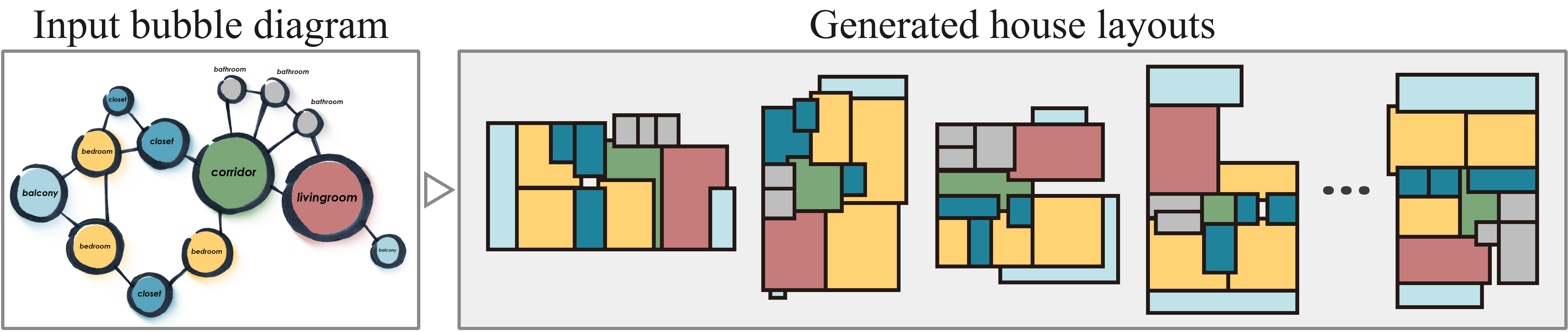}
    \caption{House-GAN is a novel graph-constrained house layout generator, built upon a relational generative adversarial network. The bubble diagram (graph) is given as an input for automatically generating multiple house layout options.}
    \label{fig:teaser}
    \vspace{-1cm}
\end{figure}

\begin{abstract}
This paper proposes a novel graph-constrained generative adversarial network, whose generator and discriminator are built upon relational architecture.
%, in particular, convolutional message passing neural networks.
The main idea is to encode the constraint into the graph structure of its relational networks. We have demonstrated the proposed architecture for a new house layout generation problem, whose task is to take an architectural constraint as a graph (i.e., the number and types of rooms with their spatial adjacency) and produce 
%a diverse set of realistic house layouts that are compatible with the input constraint
a set of axis-aligned bounding boxes of rooms. We measure the quality of generated house layouts with the three metrics: the realism, the diversity, and the compatibility with the input graph constraint.
%define the metrics of the problem to be the diversity, the realism, and the compatibility with the input graph constraint. 
Our qualitative and quantitative evaluations over 117,000 real floorplan images demonstrate that the proposed approach outperforms existing methods and baselines. We will publicly share all our code and data.

%generative adversarial network. The key idea is to employ relational neural architecture for the backbone and encode the constraint into the graph structure of its relational generator and discriminator. We demonstrate the architecture for an indoor space planning problem

%The paper We have trained the architecture with 120,000 real floorplans from LIFULL HOME's database. Our 

%This paper presents a graph-constrainted generative adversarial network for floorplan generation.

\end{abstract}
\section{Introduction}

%\yasu{Our contribution is graph-constrainted object layout generation. So the intro must talk about that capability by human. For instance, we are able to generate design a variety of spaces given certain constraints. The current intro does not seem to hit our real contribution. Pbrobably, the order bof intro should be
%\begin{enumerate}
    %\item Human is good at graph-constrained layout generation.
%    \item Computer vision made breakthroughs in unconstrained image generation such as human %faces. However, constrained image generation is at an early stage.
    %\item This paper tackles graph-constrained layout generation, in particular, floorplan generation from room type and connectivity information encoded as a graph.
%    \item Contribution summary...
%\end{enumerate}
%}
A house is the most important purchase one might make in life, and we all want to live in a safe, comfortable, and beautiful environment. However, designing a house that fulfills all the functional requirements with a reasonable budget is challenging. Only a small fraction of the residential building owners have enough budget to employ architects for customized house design. 

House design is an expensive and time-consuming iterative process. A standard workflow is to 1) sketch a ``bubble diagram'' illustrating the number of rooms with their types and connections; 2) produce corresponding floorplans and collect clients feedback; 3) revert to the bubble diagram for refinement, and 4) iterate. Given limited budget and time, architects and their clients often need to compromise on the design quality. Therefore, automated floorplan generation techniques are in critical demand with immense potentials in the architecture, construction, and real-estate industries.

\begin{figure}[!t]
 \centering
\includegraphics[width=\linewidth]{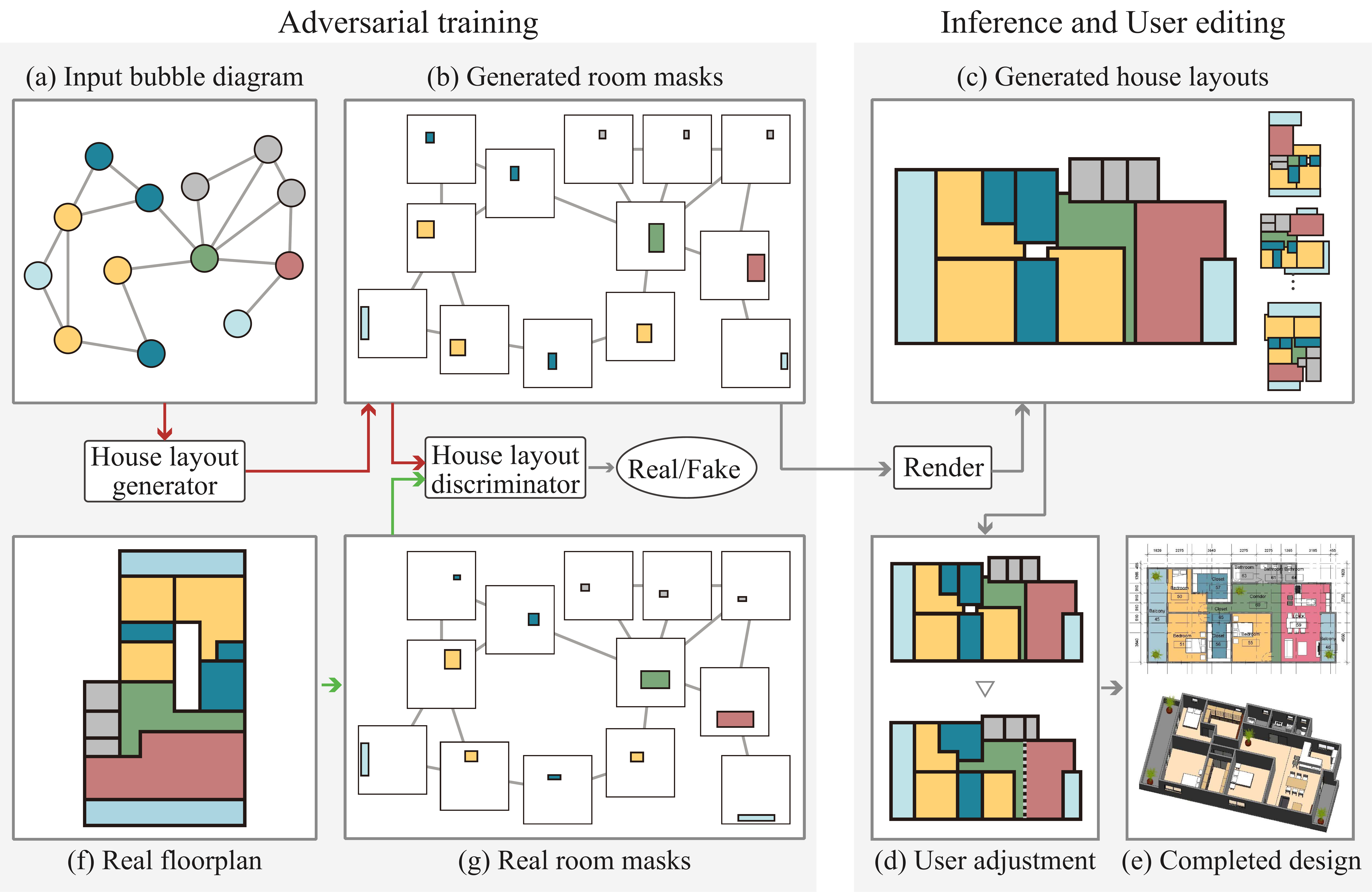}
\caption{Floorplan designing workflow with House-GAN. 
The input to the system is a bubble diagram encoding high-level architectural constraints. House-GAN learns to generate a diverse set of realistic house layouts under the bubble diagram constraint. Architects convert a layout into a real floorplan.
%System overview. Talk about input is bubble diagram. The generator produces pixel-wise room segmentation masks, where simple heuristics are used to generate vector-graphics floorplans. 
%\yasu{This figure needs polishing. As we received feedback, the problem is that the output floorplan and the input floorplan looks the same. Our output should look worse. The input should not be exactly the same floorplan and should have a different layout. Maybe, even a bubble diagram sould be different. Overall, style is also not too great (how fonts are used at the top of rounded squares. They do not look like a good title label... Just make it beautiful}}
}
%\yasu{A few comments: The figure looks a bit too scarse. We can likely compress or decrase the overall size somehow. Simply making all the figures (bubble diagram, layout, floorplan) smaller is the quickest. Use "House layout" instead of "room segments". We have a house-layout discriminator. So, the input must be either generated house layout or true (or GT) house layout. We do not use the term "room segments" in the text. Also, Floorplan should be Generated floorplan at the top right, and GT floorplan for the bottom left? Also, we have the comment during seminar on this. Our generated house layout don't have rectangular segmentations. We should change the shape to non rectangular. We do rectangular fitting in the rendering process. Also, maybe use a different floorplan example.}
 \label{figure:system}
\end{figure} 

This paper proposes a novel house layout generation problem, whose task is to take a bubble diagram as an input, and generates a diverse set of realistic and compatible house layouts (See Fig.~\ref{fig:teaser}). 
%studies a novel constrained generative model that takes a bubble diagram as an input, and generates a diverse set of realistic house layouts that are compatible with the bubble diagram. More specifically,
A bubble diagram is represented as a graph where 1) nodes encode rooms with their room types and 2) edges encode their spatial adjacency.
%(e.g., an edge exists when two rooms are adjacent).
A house layout is represented as a set of axis-aligned bounding boxes of rooms (See Fig.~\ref{figure:system}).

%in particular, real floorplan layout generation constrained on room types and connectivity between rooms encoded as a graph. We define an input graph constraint to be composed of a set of nodes corresponding to rooms and a set of edges corresponding to pairwise room adjacency (i.e. apart within a certain threshold). Some challenges in this task include precisely placing rooms in the layout, satisfying pairwise room adjacency and generating diverse layouts given an input graph constraint. 

Generative models have seen a breakthrough in Computer Vision with the emergence of generative adversarial networks (GANs)~\cite{goodfellow2014generative}, capable of producing realistic human faces~\cite{karras2019analyzing}, street-side images~\cite{kwon2019predicting}.
%\yasu{I clearly remember someone did at cvpr/iccv. Cannot find from a notable people, and  am just using it for now. May be better to replace with a better one}, or object layouts in a scene~\cite{li2019layoutgan,johnson2018image}.
%\yasu{I made up, which may be false. Make 3 examples of GAN generation here.}
%
GAN has also proven effective for constrained image generation.
%, on the other hand, is still at an early stage of research.
Image-to-image translation has been an active area of research, where an image is given as a constraint (e.g., generating a zebra image with the same pose from a horse image)~\cite{zhu2017unpaired,isola2017image,choi2018stargan}.
%\yasu{Add one more sentence for another typical constrained GAN task, while explicitly specifying what the input is. Maybe use a classification label for an input?}
Realistic scene images are generated given object bounding boxes and placements as the constraint~\cite{ashual2019specifying}.

%realistic human faces~\cite{this-person-does-not-exist-dot-com}, 
%
%This paper studies a new family of constrained generative models that takes a graph as a constraint for GAN.
%More specifically, we seek to generate a diverse set of realistic house layouts that are compatible with an input bubble diagram. A bubble diagram is represented as a graph, where 1) nodes encode rooms with their room types and 2) edges encode their spatial adjacency (e.g., an edge exists when two rooms are adjacent). A house layout is represented as room-wise segmentation masks.
%
%broadens the computational capability of GAN by allowing a graph as an input constraint. More concretely,
%Constrained GAN allows input constraints besides the noise vector in the GAN framework. \yasu{Put 1 or 2 examples of constraints GAN whose inputs are simple.} This paper explores a new family of constrainted GAN, which takes a graph as an input constraint.
%For the task of floorplan generation, the number of rooms with their room types and connections is a popular input constraint (e.g., one living-room, one kitchen, and 3 bed-rooms, where the bedrooms should not be connected to the kitchen). 
%
%this paper proposes a novel GAN framework that takes a bubble diagram represented as a graph, then generates a diverse set of realistic house layouts

The house layout generation poses a new challenge: The graph is enforced as a constraint. We present a novel generative model called House-GAN that employs relational generator and discriminator, where the constraint is encoded into the graph structure of their relational neural networks.
More specifically, we employ convolutional message passing neural networks (Conv-MPN)~\cite{conv_mpn}, which differs from graph convolutional networks (GCNs)~\cite{johnson2018image,ashual2019specifying} in that
%the way message are carried out across graph nodes, allowing this pipeline to be fully functional without any prior location constraint, generating high-quality and diverse layouts for a given input graph. 
1) a node represents a room as a feature volume in the design space (as opposed to a 1D latent vector),  and 2) convolutions update features in the design space (as opposed to multi-layer perceptron). 
The architecture enables more effective higher-order reasoning for composing layouts and validating adjacency constraints.

Our qualitative and quantitative evaluations over 117,000 real floorplan images demonstrate that House-GAN is capable of producing more diverse sets of more realistic floorplans that are compatible with the bubble diagram than the other competing methods. We will publicly share all our code and data.
%which are converted into vector graphics images by the method vectorized utilizing the method proposed in \cite{liu2017raster} and converted into a set of room binary masks and floorplan graphs for training and testing the layout generation experiments performed in this work. We summarize our contributions as follows. 

%\begin{enumerate}
%    \item Unlike existing methods, we propose an end-to-end layout generation pipeline constrained solely on graph inputs without any other prior information. %    \item Our approach provides superior layout quality in terms of meeting the given constraints and diversity of layouts in comparison to competing methods. 
%    \item ... 
%\end{enumerate}

% \nelson{talk about limitations of current procedural approaches for this problem}\\
% \nelson{talk about limitations of current VAE and GAN methods}\\
% \nelson{present our method and data}\\
% \nelson{list our contributions}\\
% End-to-end learnable pipeline that allows layout generation constrained on graphs using GAN loss

% Skeleton:
% 1. Automatic design of floor plans from bubble diagrams
%     1.1 Easily editable format in both ends (diagrams and vector)
    
% 2. Layout Generation 

%     2.1 Previous approaches are procedural and not end-to-end learnable \cite{merrell2010computer}
%     2.2 Generated layouts proposes multiple layouts for a single graph.
%     2.3 Our model learns how to arrange rooms considering the given relationships.
    
% 3. Contributions

%     3.1 End-to-end learnable pipeline that allows layout generation constrained on graphs using GAN loss

\section{Related work}

\mysubsubsection{Procedural layout generation} 
Layout composition has been an active area of research in various domains, including
%     Throughout the literature numerous works have proposed procedural approaches for composing layouts in various applications such as 
architectural layouts~\cite{harada1995interactive,bao2013generating,muller2006procedural,peng2014computing}, game-level design ~\cite{hendrikx2013procedural,ma2014game} and others. In particular, Peng \etal~\cite{peng2014computing} takes a set of deformable room templates and tiles arbitrarily shaped domains subject to the accessibility and aesthetics constraints.
%proposes an energy-based method for tiling arbitrarily shaped domains, given a set of deformable room templates subject to the 
%in order to generate ``water-tight" layouts meeting
%accessibility and aesthetics constraints.
Ma \etal~\cite{ma2014game} 
%proposes an energy based method for automatically 
generates diverse game-level layouts, given a set of 2D polygonal ``building blocks" and their connectivity constraints as a graph.
%and an input planar graph encoding block-to-block connectivity. 
These methods are more traditional and based on hand-crafted energy minimization. Our approach exploits powerful data-driven techniques (i.e., deep neural architecture) for more robustness.
%without data-driven techniques, while our approach exploits powerful deep neural architecture.
%These methods are not data-driven and utilizes non-learnable methods for satisfying constraints such as Integer Programming (IP). \yasu{What is the problem of these methods?}

\mysubsubsection{Data-driven space planning} 
Data-driven sequential generative methods have been proposed for indoor scene synthesis by Wuang \etal~\cite{wang2018deep} and Ritchie \etal~\cite{ritchie2019fast}, indoor plan generation by Wu \etal~\cite{wu2019data}, and outdoor scene generation by Jyothi \etal~\cite{jyothi2019layoutvae}.
%and a non-sequential method by Li \etal~\cite{li2019layoutgan}. 
In particular, Wu \etal~\cite{wu2019data} proposes a data-driven method for automatic floorplan generation for residential houses from a building footprint. The method starts from the living-room and sequentially adds rooms via an encoder-decoder network, followed by a final post-processing for the vectorization.
%always assuming living-room as the starting point, utilizes an encoder-decoder network to estimate walls and applies a post-processing for outputting the final vector format.
Jyothi \etal~\cite{jyothi2019layoutvae} proposes a variational autoenconder (VAE),
%based method called LayoutVAE,
which iteratively predicts a diverse yet plausible counts and sets of bounding boxes, given a set of object labels as input. 
Li \etal~\cite{li2019layoutgan} proposes a non-sequential adversarial generative method called LayoutGAN, which has a self-attention mechanism in the generator and a wireframe renderer in the discriminator.
These methods produce impressive results but cannot take a graph as an input constraint.

%which performs high-order reasoning on composing layouts in a non-sequential fashion, achieved through their self-attention mechanism in the generator and wireframe renderering discriminator.
%LayoutGAN diverges from our method when operating on bounding boxes coordinates and performing unconstrained layout predictions.
     
\mysubsubsection{Graph-constrained layout generation}
%More closely related to this paper, this line of research on
Graph-constrained layout generation has also been a focus of research.
%carried out for indoor scene synthesis by Wang \etal ~\cite{wang2019planit}, for house layout generation by
%on indoor scene synthesis, by 
%Merrell \etal~\cite{merrell2010computer}, and realistic image synthesis from scene-graphs by Jonhson \etal~\cite{johnson2018image} and Ashual \etal~\cite{ashual2019specifying}.
%
%
Wang \etal ~\cite{wang2019planit} plans an indoor scene as a relation graph and iteratively inserts a 3D model at each node via convolutional neural network (CNN) guided search.
%the PlanIT framework, which consists of planning an indoor scene via a relation-graph and instantiating it by interatively inserting elements in a 3D model for each graph node via convolutional network guided search.
%
%
Merrel \etal~\cite{merrell2010computer} utilizes Bayesian Networks for retrieving candidate bubble diagrams, given high-level conditions such as the number of rooms with room types and approximate square footage. These bubble diagrams are later converted to floorplans using the Metropolis algorithm.
Jonhson \etal~\cite{johnson2018image} and Ashual \etal~\cite{ashual2019specifying} aim to generate image layouts and synthesize realistic images from input scene-graphs via GCNs. 
%While these methods ultimately attempts to synthesize real images, our method specializes on generating realistic layouts without any location or attribute priors such as in Ashual \etal~\cite{ashual2019specifying}. Furthermore, we exploit spatial information through Conv-MPN~\cite{conv_mpn} on feature volumes instead of operating on 1d feature vector as in GCNs.
Our innovation is a novel relational generative adversarial network, where the input constraint is encoded into the graph structure of the relational generator and discriminator. 
%Our approach is non-sequential and generates the arrangement of all the rooms simultaneously that are more realistic, more diverse, and more compatible with the input constraint.
The qualitative and quantitative evaluations demonstrate the effectiveness of our approach over all the competing methods.

% The aforementioned layout generation research learns the distribution from the training data. However, there are times when human designers what to specify the relations as constraints. Both \cite{johnson2018image} and \cite{ashual2019specifying} apply a graph neural network to take in graph input and generate segmentation masks and bounding boxes for each object. A scene layout is created by aggregating over all objects and then converted to the output image. \cite{ashual2019specifying} differs from \cite{johnson2018image} in adding appearance information, class-wise aggregation for scene layout, and stochastic segmentation mask prediction. To generate a realistic image that is compatible with the input graph, not only the bounding-boxes need to follow the spatial relations, the appearance of each object also need to match the inter-object relations. For instance, a man besides a kite does not guarantee the man is looking at the kite.
    
    % A similar application work, \cite{zhou2019scenegraphnet} uses graph neural network to predict object class probability given a query location in a scene represented by a relation graph.

\section{Graph-constrained house layout generation problem} \label{section:problem}

% \yasu{Can we formally define this type of a problem? Given a set of graphic elements $V$, their pairwise relationships $E$, and a noise-vector $z$, the task is to infer a graphic layout of the elements, that is a segmentation mask per element. (can be bounding boxes)}

% \yasu{If we have a section of problem definition, we need to introduce metrics here. Data may move to results section if things get too long here. Maybe no subsections and use paragraphs to write here (dataset, task/problem, metrics)}

We seek to generate a diverse set of realistic house layouts, compatible with a bubble diagram. The section explains our dataset, metrics, and limitations.
%This paper proposes a graph-constrained house layout generation problem, whose task 
%(the starting point of any floorplan design).

\begin{figure}[!t]
     \centering
     \includegraphics[width=\linewidth]{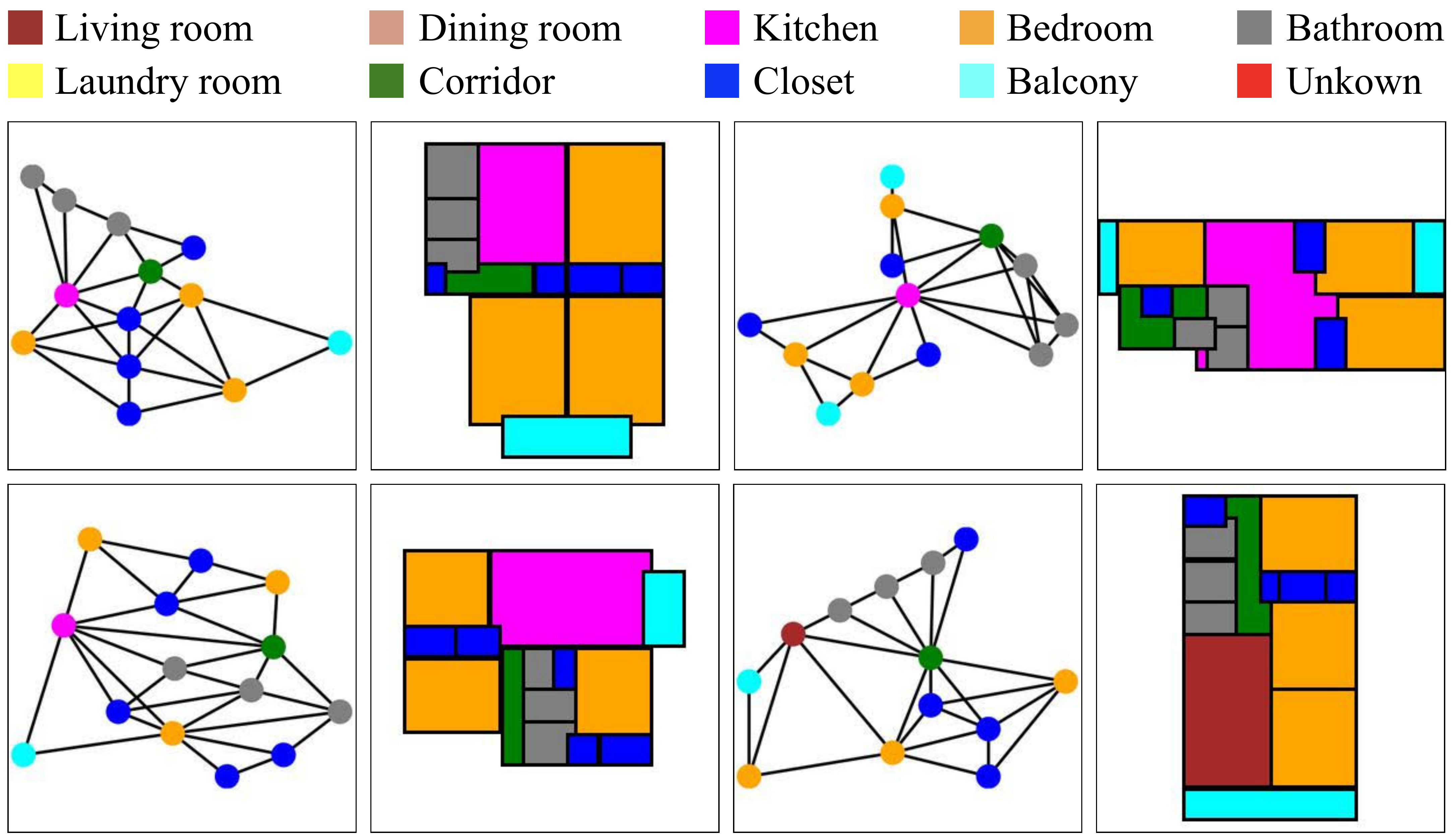}
     \caption{Sample bubble diagrams and house layouts in our dataset.}
\label{figure:gt_samples}
\end{figure}

\mysubsubsection{Dataset}
LIFULL HOME's database offers five million real floorplans, from which we retrieved 117,587~\cite{lifull} and rescaled uniformly to fit inside the $256\times 256$ resolution (See Table~\ref{tab:stats}).
%Floorplans are rescaled uniformly to fit inside $256\times 256$ square.
The database does not contain bubble diagrams. We used the floorplan vectorization algorithm~\cite{liu2017raster} to generate the vector-graphics format, which is converted into bubble diagrams.  A bubble diagram is a graph, where a node is a room with a room type as its property.~\footnote{Room types are ``living room'', ``kitchen'', ``bedroom'', ``bathroom'', ``closet'', ``balcony'', ``corridor'', ``dining room'', ``laundry room'', or ``unkown''.} Two rooms are connected if 
%the corresponding bounding boxes are within \yasu{X} pixels in the original floorplan).
the Manhattan distance between the bounding boxes is less than 8 pixels.
%\nelson{maybe mention that we consider adjacent boxes inside to each other.}
%then generated bubble-grams by the following heuristics: 1) Fit an axis aligned bounding box to each room; 2) Make a node for each room; and 3) Add an edge if the bounding boxes of the two rooms are within \yasu{X} pixels.
%
%
%\begin{figure}
%\centering
        %\caption{Dataset histogram.}
    %\label{fig:dataset_hist}
%\end{figure}
An output house layout is axis-aligned bounding boxes (See Fig.~\ref{figure:gt_samples}).

    % Model & \multicolumn{5}{c}{Edit distance} & \multicolumn{5}{c}{FID} \\
    % \cmidrule(lr){2-6}\cmidrule(lr){7-11} &
    % 1-3 & 4-6 & 7-9 & 10-12 & 13+ & 1-3 & 4-6 & 7-9 & 10-12 & 13+ \\

% %%%%%%%%%%%%%%%%%%%%% Dataset stats %%%%%%%%%%%%%%%%%%%%%%%
\begingroup
\renewcommand{\arraystretch}{1.1}
\begin{table}[!t]
    \caption{We divide the samples into five groups based on the room counts (1-3, 4-6, 7-9, 10-12, and 13+). The second row shows the numbers of samples. The remaining rows show the average number of rooms (left) and the average number of edge connections per room (right). 
    %For example, among floorplans whose room-counts are in the range [10-12], the average number of bathrooms is 3.0 and a bathroom is connected to 3.3 rooms on the average.
    The room counts are small for living-rooms, which are often interchangeable with dining-rooms and kitchens in the Japanese real-estate market.
    %Note that our data comes from the Japanese real-estate market living room is
    %. We divide the samples into five groups based on the room-counts (1-3, 4-6, 7-9, 10-12, and 13+), and report the statistics.
    %
    %Shows the average number of room types, average number of edges for each room type and number of samples in each graph group (i.e. 1-3, 4-6, 7-9, 10-12, and 13+) and in the entire dataset (``All").}
    }
    \label{tab:stats}
    \centering
    \begin{tabular}{lC{1.5cm}C{1.5cm}C{1.5cm}C{1.5cm}C{1.5cm}C{1.5cm}}
    \toprule
    \cmidrule(lr){2-7}  &
    1-3 & 4-6 & 7-9 & 10-12 & 13+ & All\\
    \midrule
    (\# of samples) & 7,393 & 28,170 & 42,635 & 30,625 & 8,764 & 117,587\\ 
    \midrule
%    & \multicolumn{6}{c}{(\# of rooms) / (\# of edge connections per room)} \\
    %   & \multicolumn{2}{c}\\
    % \cmidrule(lr){2-6}
    Living Room  &  0.0/1.2  &  0.0/3.1  &  0.1/4.8  &  0.3/5.6  &  0.3/5.9  & 0.1/5.1\\
    Kitchen      &  0.6/1.3  &  1.0/3.3  &  1.2/4.5  &  1.1/5.4  &  1.3/5.3  & 1.1/4.4\\
    Bedroom      &  0.4/1.3  &  0.8/2.8  &  1.3/3.5  &  2.0/3.9  &  2.8/4.1  & 1.4/3.6\\
    Bathroom     &  0.7/1.2  &  1.6/2.4  &  2.6/2.9  &  3.0/3.3  &  3.4/3.5  & 2.4/3.0\\
    Closet       &  0.3/1.2  &  1.0/2.2  &  1.6/2.6  &  2.4/3.1  &  3.6/3.2  & 1.7/2.8\\
    Balcony      &  0.2/0.9  &  0.6/1.2  &  0.9/1.5  &  1.0/1.9  &  1.3/2.0  & 0.8/1.6\\
    Corridor     &  0.1/1.1  &  0.1/2.6  &  0.4/3.7  &  1.0/4.6  &  1.4/5.0  & 0.5/4.3\\
    Dining Room  &  0.0/1.5  &  0.0/3.0  &  0.0/3.6  &  0.0/3.2  &  0.0/1.9  & 0.0/2.9\\
    Laundry Room &  0.0/0.0  &  0.0/0.0  &  0.0/0.0  &  0.0/0.0  &  0.0/0.0  & 0.0/0.0\\
    Unknown      &  0.0/0.0  &  0.0/0.0  &  0.0/0.0  &  0.0/0.0  &  0.0/0.0  & 0.0/0.0\\
    \bottomrule
\end{tabular}
\end{table}
\endgroup

\mysubsubsection{Metrics}
We divide the samples into five groups based on the number of rooms: (1-3, 4-6, 7-9, 10-12, and 13+).
%For the generation of house-layouts in a group, we train a model using samples in the other groups
To test the generalization capability, we conduct k-fold validation (k=5): When generating layouts in a group, we train a model while excluding samples in the same group so that a method cannot simply memorize.
%\nelson{I think one of the selling points is that our method generalize better in k-fold cross validation, should we explain why this is good/useful?}
%so that a method cannot just remember results.
%we perform group-wise cross validation: For example, when generating a layout with 8 rooms, we train a model while excluding samples whose room sizes are in [7-9].
%: Whenb generating layouts in a group, we exclude 
%while excluding samples in the group for testing the generalization capability. 
At test time, we randomly pick a house layout and generate X samples. $X=10$ for measuring the realism and diversity, and $X=1$ for measuring the compatibility whose evaluation is computationally expensive.
%\nelson{this number is only used for realism and diversity, for compatibility we just use 1, too expensive to compute} from its bubble diagram.
%\yasu{Do we change how we run things at test-time depending on the metrics? If we only change for one metric, we can explain that as an exception later. here we can describe the most common test-time scenario.}. \nelson{For computing the graph similarity we just generate 1 sample per bubble diagram for 5k samples in the removed group (that's already quite costly computationally).}
%
%There are three metrics.
%(See Sect.~\ref{sect:metrics} for the details).
%measure the performance: realism, diversity, and compatibility.

\noindent $\bullet$
The realism is measured by an average user rating. We present a generated house layout against a ground-truth or another method. A subject puts one of the four ratings: better (+1), worse (-1), equally-good (+1), or equally-bad (-1).
%+1/-1 (better), +1/+1 (equally good), -1/-1 (equally bad), or -1/+1 (worse).
%user study, in which a subject sees a pair of house-layouts then chooses more realistic sample or rates as a tie.

\noindent $\bullet$ The diversity is measured by the FID score~\cite{heusel2017gans} with the rasterized layout images. We rasterize a layout by 1) Setting the background to white; 
2) Sorting the rooms in the decreasing order of the areas; and 3) Painting each room with a color based on its room type (e.g., orange for a bedroom) as shown in Figure~\ref{figure:gt_samples}. 

\noindent $\bullet$ The compatibility with the bubble diagram is a graph editing distance~\cite{abu2015exact} between the input bubble diagram and the bubble diagram constructed from the output layout in the same way as the GT preparation above.
%Please see Section~\ref{section:raster} for the implementation details.

%used simple heuristics to 
%generate the vector-graphics floorplan 
%, our task poses new challenges in 
%Our generative task consists of 

\mysubsubsection{Assumptions}
In contrast to the real design process, we make a few restrictive assumptions to simplify the problem setting: 1) A node property does not have a room size; 2) A room shape is always a rectangle; and 3) An edge property (i.e., room adjacency) does not reflect the presence of doors.
This is the first research step in tackling the problem, where these extensions are our future work.

%We represent a floor plan layout configuration as a graph $\mathcal{G}$, where nodes correspond to rooms and edges to adjacency between rooms nodes. We assume here that any room can be represented by a single rectangular shape, thus we avoid touching the problem of learning the distribution of rooms shapes, which would increase significantly the problem complexity and focus on learning distributions of relative positioning between rectangular shapes. Two rooms are considered to be adjacent if their vertical and horizontal clear spacing is within a certain threshold. Given two bounding boxes with sizes $(w_1, h_1)$ and $(w_2, h_2)$, centered at $(x_1, y_1)$ and $(x_2, y_2)$, respectively, we compute the distance between them as follows:   
%        \begin{equation}
%            d = \max(|x_1-x_2|-\frac{w_1+w_2}{2}, |y_1-y_2|-\frac{h_1+h_2}{2})
%        \end{equation}

%Evaluation details}
%    \subsubsection{Dataset}
    
 %\subsubsection{Metrics}
 %   We show results for graph similarity edit distance and
%    \yasu{Ultimately, we want to evaluate how realistic, our room arrangement is. Maybe a user study? (I am not 100\% confident and want to discuss: We cannot generate high quality floorplan with perfect vector-graphics. Maybe, we visualize rough room location and area by applying smoothing on the shapes. We show the visualization on real examples to a users, then show one sample and ask if this is real or generated. For vision conference, this is nice, but for a ML conference, user study may not be important.}
    
    %\nelson{graph similarity and FID}

\section{House-GAN}
House-GAN is a relational generative adversarial network.
%whose specialization is the relational generator and discriminator
%that generates a set of house layouts with the bubble diagram constraint
%and is composed of a generative and a discriminative network optimized to find equilibrium in a minimax game given a certain input constraint.
The key specialization is our relational generator and discriminator,
%in particular Conv-MPN~\cite{conv-mpn},
where the input graph constraint is encoded into the graph structure of the relational networks.
In particular, we employ Conv-MPN~\cite{conv_mpn}, which differs from GCNs~\cite{ashual2019specifying,johnson2018image} in that a node stores a feature volume and convolutions update features in the design space (as opposed to a 1D latent vector space).

\subsection{House layout generator}
\begin{figure}[tb]
\centering
\includegraphics[width=\linewidth]{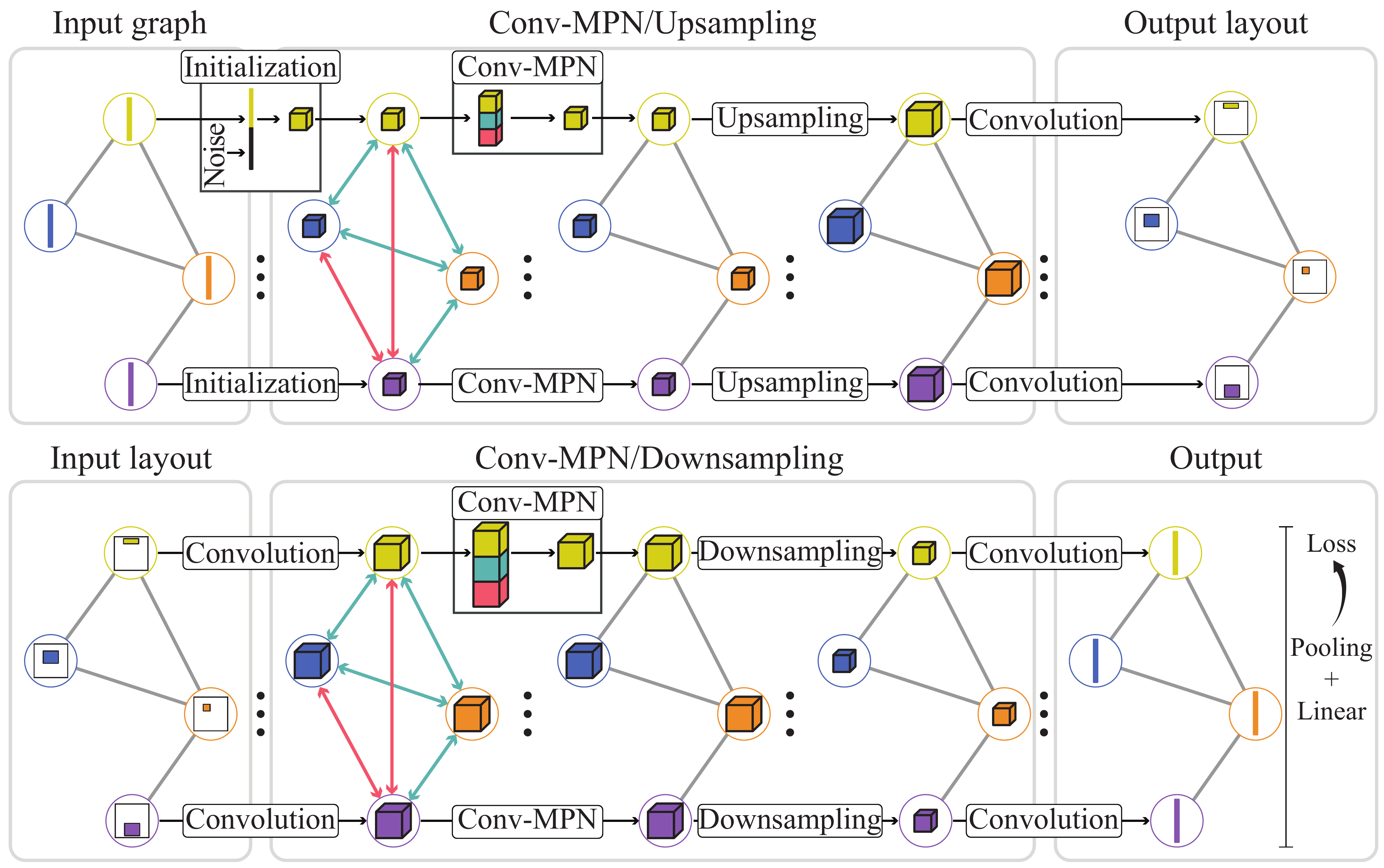}
\caption{Relational house layout generator (top) and discriminator (bottom). Conv-MPN is our backbone architecture~\cite{conv_mpn}. The input graph constraint is encoded into the graph structure of their relational networks. }
%and information is propagated through Conv-MPNs~\cite{conv_mpn}.\nelson{Maybe we should use Conv-MPN instead of CMP for consistency? CMP is used in Table~\ref{table:full_architecture}and seems more correct, maybe keep it}.
\label{fig:graphs}
\end{figure}
    
The generator takes a noise vector per room and a bubble diagram, then generates a house layout as an axis-aligned rectangle per room. The bubble diagram is represented as a graph, where a node represents a room with a room type, and an edge represents the spatial adjacency. More specifically, a rectangle should be generated for each room, and two rooms with an edge must be spatially adjacent (i.e., their Manhattan distance should be less than 8 pixels).
%the Manhattan distance of two rooms with an edge should be within 8 pixels. 
We now explain the three phases of the generation process (See Fig.~\ref{fig:graphs}). The full architectural specification is shown in Table~\ref{table:full_architecture}. 
%We will also share our code and data.
%illustrates the overall architecture.
%

\mysubsubsection{Input graph}
Given a bubble diagram, we form Conv-MPN whose relational graph structure is the same as the bubble diagram.
%: Generating a node for each room, and exchanging messages between connected rooms.
We generate a node for each room and initialize
%an input bubble diagram, where each of its room nodes
with a 128-d noise vector sampled from a normal distribution, concatenated with a 10-d room type vector $\overrightarrow{t_r}$ (one-hot encoding). $r$ is a room index.
%
%respective given room node type (i.e. 10-d vector one-hot encoded),
This results in a 138-d vector $\overrightarrow{g_r}$:
\begin{equation}
    \overrightarrow{g_r} \leftarrow \left\{ \mathbb{N}(0,1)^{128} ; \overrightarrow{t_r} \right\}.
\end{equation}
%
%Different from standard graph neural networks, 
Conv-MPN stores features as a 3D tensor in the output design space.
%represents a latent feature as a 3D feature volume in the output design space.
%, instead of a 1D vector for each node.
% a node represents a room as a feature volume in the design space (as opposed to a 1D latent vector), 
We apply a shared linear layer for expanding $\overrightarrow{g_r}$ 
%the initial 138-d vector
into a ($8\times8\times16$) feature volume $\mathbf{g}_r^{l=1}$. $(l=1)$ denotes that the feature is for the first Conv-MPN module, which will be upsampled twice to become a ($32\times32\times 16$) feature volume $\mathbf{g}_r^{l=3}$ later.
%, which will be upsampled twice.
%, which will be repeated three times. %with upsampling.
%while upsampling the feature resolution.
%by a factor of 2 every time.
%e level of the feature volume, \nelson{at a first glance not very clear to me what level of feature volume means} which will be upsampled in the next step.

\mysubsubsection{Conv-MPN/Upsampling}
%A graph of room-wise 3D feature volumes is our latent house layout representation throughout our architecture.
Conv-MPN module updates a graph of room-wise feature volumes via convolutional message passing~\cite{conv_mpn}.
%\nelson{Conv-MPN instead of CMP?}. 
More precisely, we update $\mathbf{g}_r^{l}$ by 1) concatenating a sum-pooled feature across rooms that are connected in the graph; 2) concatenating a sum-pooled feature across non-connected rooms; and 3) applying a CNN:
\begin{equation}
    \mathbf{g}_r^l \leftarrow \mbox{CNN}\left[\mathbf{g}_r^l\quad ;\quad \underset{s\in \mathbf{N}(r)}{\mbox{Pool}} \mathbf{g}_s^l \quad ; \quad \underset{s \in \overline{\mathbf{N}}(r)}{\mbox{Pool}} \mathbf{g}_s^l \right].
\end{equation}
$\mathbf{N}(r)$ and $\overline{\mathbf{N}}(r)$ denote sets of rooms that are connected and not-connected, respectively.
We upsample features by a factor of 2 using 
%We apply two rounds of upsampling plus convolutional message passing (CMP) ~\cite{conv_mpn}. The upsampling layer consists of
a transposed convolution (kernel=4, stride=2, padding=1), while maintaining the number of channels.
%The CMP layer updates each room node feature volume by pooling features over all its directly connected neighbours (sharing an edge) into a feature volume and non connected neighbours (not sharing an edge) into a second feature volume, which are concatenated along with the current node feature volume and encoded by a convolutional neural network (CNN), resulting in the updated feature volume. At this point, each graph node contains a $16\times32\times32$ feature volume,
The generator has two rounds of Conv-MPN and upsampling, making the final feature volume $\mathbf{g}_r^{l=3}$ of size ($32 \times 32 \times 16$).
%\nelson{Why l=3 if we have only 2 conv-mpn?}.

\mysubsubsection{Output layout}
A shared three-layer CNN converts a feature volume into a
%probabilistic \nelson{Not really. Last nonlinearity is tanh [-1,1]} 
room segmentation mask of size ($32\times32\times1$).
This graph of segmentation masks will be passed to the discriminator during training. At test time, the room mask (an output of tanh function with the range [-1, 1]) is thresholded at 0.0, and we fit the tightest axis-aligned rectangle for each room to generate the house layout.
%for the generation of the house layout, we threshold the room segment by 0.0 \nelson{we threshold at 0.0 (mid range value, [-1, 1] is the output range)}, then fit the tightest axis-aligned rectangle for each room.

\subsection{House layout discriminator}
The discriminator performs a sequence of operations in the reverse order.
%The discriminator performs a sequence of symmetric operations in comparison to the generator. 
The input is a graph of room segmentation masks
%\nelson{we normalize binary masks [0,1] to [-1,1] for samples/for generated not necessary (output is already in this range)} 
either from the generator (before rectangle fitting) or a real floorplan (1.0 for foreground and -1.0 for background). A segmentation mask is of size $32\times32\times1$.
To associate the room type information, we take a 10-d room type vector, apply a linear layer to expand to 8192-d, then reshape to a $(32\times32\times8)$ tensor, which is concatenated to the segmentation mask.
%\nelson{"expand it (1x8192) then reshape (32x32x8) and concatenate it with its room mask."}
%\yasu{Please write what you did here}
%simply concatenate a 10-d room type vector to the input segmentation mask at every pixel, making the input a $32\times32\times9$ tensor.
%\yasu{This does not agree with the table. We take a room type and transform to 8x32x32?}
%\nelson{We upsample type vector to ($32\times32\times8$) and concatenate with the mask.}
%
A shared three-layer CNN converts the feature into a size ($32\times 32\times 16$), followed by two rounds of Conv-MPN and downsampling. We downsample by a factor of 2 each time by a convolution layer (kernel=3, stride=2, padding=1).
%Given an input bubble diagram and a candidate house layout (either from the training set or generated set), we initialize each room node with its corresponding room mask ($1\times32\times32$). A shared three-layered CNN is applied, converting each room mask into feature volumes ($16\times32\times32$). Next, we apply two rounds of downsampling plus CMP layer. Analogously, the downsampling decreases the feature resolution by a factor of 2 and the CMP layer updates feature volumes as in the generator. 
%
Lastly, we use a three layer CNN for converting a room feature into a 128-d vector ($\overrightarrow{d_r}$). We sum-pool over all the room vectors and add a single linear layer to output a scalar $\mathbf{d}$, classifying
%\nelson{Not confident on this term as we use WGAN-GP (distance not classification)} 
ground-truth samples from generated ones. 
\begin{equation}
\mathbf{d} \leftarrow \mbox{Linear}(\underset{r}{\mbox{Pool}}\ \overrightarrow{d_r})
\end{equation}
\renewcommand{\arraystretch}{1.1}
    \begin{table*}[!h]
    \caption{House-GAN architectural specification.
    %Shows architecture details for the proposed house layout generator and discriminator in Space-GAN. 
    ``s" and ``p" denote stride and padding.
    %in convolution layers. 
    ``x", ``z" and ``t" denote the room mask, noise vector, and room type vector. ``conv\_mpn" layers have the same architecture in all occurrences. Convolution kernels and layer dimensions are specified as ($N_{in}\times N_{out} \times W \times H$) and ($W \times H \times C$).}
    %as convolution kernels and layer output dimensions, respectively.}
    \label{table:full_architecture}
    \centering
        \begin{tabular}{cccc}
        \toprule
        Architecture & Layer & Specification & Output Size\\
        \midrule
        \midrule
        %%%%%%%%%%%%%%%%%%%%%%%%%%%%
        %%%%%%% Generator %%%%%%%%%%
        %%%%%%%%%%%%%%%%%%%%%%%%%%%%
        & $concat(z, t)$ & N/A &  $1\times138$\\
        \cmidrule{2-4}
        & $linear\_reshape_1$ & $138\times1024$ &  $8\times8\times16$\\
        \cmidrule{2-4}
        & $conv\_mpn_1$ & ${\left[\begin{array}{c}16\times16\times3\times3, (s=1, p=1)\\16\times16\times3\times3, (s=1, p=1)\\16\times16\times3\times3, (s=1, p=1)\\\end{array} \right ]}$ &  $8\times8\times16$\\
        \cmidrule{2-4}
        \multirow{2}{*}{\parbox{2cm}{\centering House layout\\ generator}} & $upsample_1$ & $16\times16\times4\times4, (s=2, p=1)$ &  $16\times16\times16$\\
        \cmidrule{2-4}
        & $conv\_mpn_2$ & - &  $16\times16\times16$\\
        \cmidrule{2-4}
        & $upsample_2$ & $16\times16\times4\times4, (s=2, p=1)$ &  $32\times32\times16$\\
        \cmidrule{2-4}
        & $conv\_leaky\_relu_1$ & $16\times256\times3\times3, (s=1, p=1)$ &  $32\times32\times256$\\
        \cmidrule{2-4}
        & $conv\_leaky\_relu_2$ & $256\times128\times3\times3, (s=1, p=1)$ &  $32\times32\times128$\\
         \cmidrule{2-4}
        & $conv\_tanh_1$ & $128\times1\times3\times3, (s=1, p=1)$ &  $32\times32\times1$\\
        \midrule
        %%%%%%%%%%%%%%%%%%%%%%%%%%%%%%%%
        %%%%%%% Discriminator %%%%%%%%%%
        %%%%%%%%%%%%%%%%%%%%%%%%%%%%%%%%
        & $linear\_reshape_1(t)$ & $10\times8192$ &  $32\times32\times8$\\
        \cmidrule{2-4}
        & $concat(t, x)$ & N/A &  $32\times32\times9$\\
        \cmidrule{2-4}
        & $conv\_leaky\_relu_1$ & $9\times16\times3\times3, (s=1, p=1)$ &  $32\times32\times16$\\
        \cmidrule{2-4}
        & $conv\_leaky\_relu_2$ & $16\times16\times3\times3, (s=1, p=1)$ &  $32\times32\times16$\\
        \cmidrule{2-4}
        & $conv\_leaky\_relu_3$ & $16\times16\times3\times3, (s=1, p=1)$ &  $32\times32\times16$\\
        \cmidrule{2-4}
        & $conv\_mpn_1$ & - &  $32\times32\times16$\\
        \cmidrule{2-4}
        \multirow{2}{*}{\parbox{2cm}{\centering House layout\\ discriminator}} & $downsample_1$ &  $16\times16\times3\times3, (s=2, p=1)$ &  $16\times16\times16$\\
        \cmidrule{2-4}
        & $conv\_mpn_2$ & - &  $16\times16\times16$\\
        \cmidrule{2-4}
        & $downsample_2$ &  $16\times16\times3\times3, (s=2, p=1)$ &  $8\times8\times16$\\
        \cmidrule{2-4}
        & $conv\_leaky\_relu_1$ & $16\times256\times3\times3, (s=2, p=1)$ &  $4\times4\times256$\\
        \cmidrule{2-4}
        & $conv\_leaky\_relu_2$ & $256\times128\times3\times3, (s=2, p=1)$ &  $2\times2\times128$\\
        \cmidrule{2-4}
        & $conv\_leaky\_relu_3$ & $128\times128\times3\times3, (s=2, p=1)$ &  $1\times1\times128$\\
        \cmidrule{2-4}
        & $pool\_reshape\_linear_1$ & $128\times1$ &  $1$\\
        \bottomrule
        \end{tabular}
    \end{table*}

We use the WGAN-GP~\cite{gulrajani2017improved} loss with gradient penalty set to 10. We compute the gradient penalty as proposed by Gulrajani \etal \cite{gulrajani2017improved}: Linearly and uniformly interpolating room segmentation masks pixel-wise between real samples and generated ones, while fixing the relational graph structure.

%maintaining the fixed
%the input constraints when uniformly sampling along straight lines between pairs of points sampled from the real data and the generator distributions.
%fixing the relational graph structure and interpolating room segmentation masks pixel-wise between real samples and generated ones

%we apply a second three-layered CNN for converting room feature vectors into 128-d each followed by a pooling layer to combine all room vectors into a single global vector (128-d), which is later converted into a scalar by a single linear layer. The final scalar value is utilized for optimizing the adversarial loss.  

\section{Implementation Details}
We have implemented the proposed architecture in PyTorch and utilized a workstation with dual Xeon CPUs and dual NVIDIA Titan RTX GPUs. Our model adopts WGAN-GP~\cite{gulrajani2017improved} with ADAM optimizer ($b_1=0.5$, $b_2=0.999$) and is trained for 200k iterations. The learning rates of the generator and the discriminator are $0.0001$, respectively. The batch size is $32$.
%, and the noise vector size is $128$.
We set the number of critics to 1 and use leaky-ReLUs ($\alpha$=0.1) for all non-linearities 
%in our architecture 
except for the last one in the generator where we use hyperbolic tangent. We tried but do not use spectral normalization in the convolution layers and a per-room discriminator (before final sum-pooling), which did not lead to significant improvements.
%\nelson{should this go here?}
%\yasu{Is this correct? We did not use these techniques at the end?} \nelson{we did not use this in the end, but I tried these thing and did not help. Seems obvious things to try i feel if we don't comment reviewers will ask.}
%
%
See Table~\ref{table:full_architecture} for the full architectural specification.
%We refer the remaining implementation details and full architectural specification to the supplementary document. \nelson{Table~\ref{table:full_architecture}}

%\begingroup

%\endgroup

%\multirow{2}{*}{\parbox{2.5cm}{\centering House layout\\ discriminator}}
\section{Experimental Results}
%%%%%%%%%%%%%%%%%%%%% Main results - Edit distance %%%%%%%%%%%%%%%%%%%%%%%
\begingroup
\renewcommand{\arraystretch}{1.1}
\begin{table}[!t]
\caption{The main quantitative evaluations. Realism is measured by a user study with graduate students and professional architects. Diversity is measured by the FID scores. Compatibility is measured by the graph edit distance. $(\uparrow)$ and $(\downarrow)$ indicate the-higher-the-better and the-lower-the-better metrics, respectively.
%and FID values for target graphs of sizes: 1-3, 4-6, 7-9, 10-12 and 13+. 
We compare House-GAN against two baselines and two competing methods.
The \textcolor{cyan}{cyan}, \textcolor{orange}{orange}, and \textcolor{magenta}{magenta} colors indicate the first, the second, and third best results, respectively.}
\label{tab:main_results}
\centering
\begin{tabular}{lccccccccccc}
\toprule
    & \multicolumn{1}{c}{Realism $(\uparrow)$} & \multicolumn{5}{c}{Diversity $(\downarrow)$} & \multicolumn{5}{c}{Compatibility $(\downarrow)$} \\
    \cmidrule(lr){2-2}\cmidrule(lr){3-7}\cmidrule(lr){8-12} Model &
    All groups & 1-3 & 4-6 & 7-9 & 10-12 & 13+ & 1-3 & 4-6 & 7-9 & 10-12 & 13+ \\
    \midrule
    CNN-only & \textcolor{magenta}{-0.49} & \textcolor{cyan}{13.2} & \textcolor{magenta}{26.6} & \textcolor{magenta}{43.6} & \textcolor{magenta}{54.6} & \textcolor{magenta}{90.0} & \textcolor{magenta}{0.4} & 3.1 & 8.1 & \textcolor{magenta}{15.8} & 34.7 \\
    GCN &  \textcolor{orange}{0.11} & \textcolor{magenta}{18.6} & \textcolor{orange}{17.0} & \textcolor{orange}{18.1} & \textcolor{orange}{22.7} & \textcolor{orange}{31.5} & \textcolor{cyan}{0.1} & \textcolor{cyan}{0.8} & \textcolor{cyan}{2.3} & \textcolor{cyan}{3.2} & \textcolor{cyan}{3.7} \\
    Ashual \etal ~\cite{ashual2019specifying} &  -0.61 & 64.0 & 92.2 & 87.6 & 122.8 & 149.9 & \textcolor{orange}{0.2} & 2.7 & 6.2 & 19.2 & 36.0 \\
    Johnson \etal ~\cite{johnson2018image} & -0.62 & 69.8 & 86.9 & 80.1 & 117.5 & 123.2 & \textcolor{orange}{0.2} & \textcolor{magenta}{2.6} & \textcolor{magenta}{5.2} & 17.5 & \textcolor{magenta}{29.3}\\
    House-GAN [Ours] & \textcolor{cyan}{0.15} & \textcolor{orange}{13.6} & \textcolor{cyan}{9.4} & \textcolor{cyan}{14.4} & \textcolor{cyan}{11.6} & \textcolor{cyan}{20.1} & \textcolor{cyan}{0.1} & \textcolor{orange}{1.1} & \textcolor{orange}{2.9} & \textcolor{orange}{3.9} & \textcolor{orange}{10.8}\\
    \bottomrule
\end{tabular}
\end{table}
\endgroup

%118,012 bubble-gram/floorplan samples are divided into five groups based on the number of rooms (Also see Sect.~\ref{section:problem}). Group-wise cross validation is the 

Realism, diversity, and compatibility metrics evaluate the performance of the proposed system
%against the four competing methods. 
%examples for the generation.
against the two baselines and the two competing methods.
%from the layout generation literature utilized for comparison. 
We first introduce these methods, while referring to the supplementary document for the full architectural specification.
%cd ..We refer their detailed architecture to the supplementary document.
%

%\mysubsubsection{Baselines} 
\vspace{0.1cm}

\noindent $\bullet$ {\bf CNN-only}: We encode the bubble diagram into a fixed dimensional vector by assuming at most 40 rooms and sorting the rooms based on the room-center x-coordinate in the corresponding floorplan.
%(sort by the x-coordinate, then y-coordinate in a case of a tie). 
To be precise, we concatenate a 128-d noise vector, a 10-d room type vector for 40 rooms, and 780=$40\choose 2$ dimensional vector indicating the room connectivity, resulting in a 1308-d vector.
%room adjacency information
%. We assume 40 rooms, making the input a 528-d vector with
We pad zeros for missing rooms.
We convert a vector into a feature volume and apply two rounds of upsampling and CNN to produce room masks as a ($32\times32\times 40$) feature volume.
The discriminator takes the room masks,
%\nelson{all the rooms masks (this is not per room operation) we have one $32\times32\times8$ for the entire layout},
concatenates the room type and connectivity information (i.e. 1308-d vector) represented as a ($32\times32\times8$) feature volume and performs an inverse operation of the generator. 
%\nelson{To be precise we take room type vectors plus graph (1x1,038) \textrightarrow linear layer (1x1x8192) \textrightarrow reshape (32x32x8) then concatenate (32x32x48)}
%\yasu{Is the input 32x32x360 ?}\nelson{we have a 1d vector for types (1x400) and 1d vector for edges (40 choose 2 = 1x780) concatenate everything (1x1180) then apply linear layer (1x8,192) - reshape (32x32x8) - concatenate with masks input becomes (32x32x48) - then CNN}
%, just like our architecture.
%vector assuming a fixed ordering on the room boxes centers (i.e. top-left to bottom-right), concatenate with a noise vector (128-d) and output a fixed number of room segments (i.e. 40). The discriminator inputs a fixed number of room segments and converts into a single scalar using convolution layers. 
%We generate a fixed number of room segments (i.e. 40) with padding for a given bubble-gram using only convolution layers. 

\noindent $\bullet$ {\bf GCN}: The generator takes a 128-d noise vector concatenated with a 10-d room type vector per room. After 2 rounds of message passing as 1d vectors by GCN, a shared CNN module decodes the vector into a mask.
%\nelson{input noise is (1x128) concatenate with its type become (1x138) apply a linear layer to make it (1x128) apply GCN, the output is (1x128) apply a second linear layer output is (1x1,024) reshape it (8x8x16) apply CNN-upsample util it becomes (32x32x1)}
%The vector is , which are decoded using a shared CNN after a few rounds of message passing operating on 1-d vectors. 
The discriminator merges the room segmentation and type into a feature volume as in House-GAN. A shared CNN encoder converts it into a feature vector,
%takes a room type vector, expands it using a linear layer ($32\times32\times8$) and concatenate it with its room mask ($32\times32\times1$). The input feature volume ($32\times32\times9$) is converted into a feature vector by a shared CNN encoder, 
followed by 2 rounds of message passing,  sum-pooling, and a linear layer to produce a scalar. 
%\nelson{Discriminator: input is (32x32x1) type is (1x10), apply linear layer in type and reshape make it (32x32x8) concatenate with mask (32x32x9) apply CNN and make it 1d (1x128) apply GCN and then make it using MLP (1x1)}
%each room segment into a 1-d vector using a shared CNN and performs message passing followed by sum-pooling and a linear layer, outputting a single scalar.    
        
\noindent $\bullet$
%\mysubsubsection{Competing methods}
{\bf Ashual \etal~\cite{ashual2019specifying}} and {\bf Johnson \etal~\cite{johnson2018image}}: 
After converting our bubble diagram and floorplan data into their representation, we use their official code to train the models
%Their input is bounding boxes, room types, edges, masks 
%\yasu{Comment on what are the inputs? Do we have to do some data format conversion like the other methods above? Nothing is written and it seems helpful to explain something} \nelson{Their method is a bit complicated: In short they input bounding boxes, types, edges, masks and final images, have losses in boxes, masks and in the final image Ashual \etal also input location and object attribute priors.}
%The methods were trained on our data 
with two minor adaptations: 1) we limit scene-graphs to contain only two types of connections: ``adjacent" and ``not adjacent"; 2) we provide the rendered bounding boxes filled with their corresponding color during training. 
%We utilize the outputted set of 
%Bounding boxes become the estimated house layout during testing.

% The method was trained without object and location attributes, they were both zeroed as our task requires layout estimation without any object and location priors. Both competing methods were trained on our data by making two adaptations from the official code: 1) we assume scene-graphs to contain only two types of connections: ``adjacent" and ``not adjacent"; 2) we provide the rendered house layout image for training these methods. We utilize the outputted set of bounding boxes as the estimated house layout for evaluation.

%\noindent $\bullet$: We make the same adaptations as for Ashual \etal~\cite{ashual2019specifying} in the official code.
%for training this method on our data and we utilize their outputted set of bounding boxes as the house layout predictions during testing phase. 

\vspace{0.2cm}
Table~\ref{tab:main_results} shows our main results. As explained in Sect.~\ref{section:problem}, we divide 117,587 samples into 5 groups based on their room counts. For the generation of layouts in each group, we exclude samples in the same group from the training so that methods cannot simply memorize layouts.
%comparing our approach House-GAN against the two baselines (CNN-only and GCN) and competing state-of-the-arts: 1) Johnson \etal~\cite{johnson2018image}; 2) Ashual \etal~\cite{ashual2019specifying}. In short, 
House-GAN outperforms the competing methods and the baselines in all the metrics, except for the compatibility against GCN with a small margin.
%CNN-only baseline by a large margin in all the metrics. House-GAN also outperforms
%
%The GCN baseline is on par in realism and better in compatibility with a small margin.
%outperforms House-GAN on larger houses by a small margin on the compatibility score, 
%in particular for graphs of size (13+), 
%where the disparity is more significant. 
%The main difference is the diversity, where GCN baseline seems to rather copy and recycle a few ground-truth samples from the training time with minor modifications.
%one realistic sample with minor modifications
%However, House-GAN beats the GCN baseline by a large margin on FID score (diversity) as shown in Table~\ref{tab:main_results} and in our qualitatively results shown in Fig.~\ref{figure:diversity}.
We now discuss each of the three metrics in more detail with more qualitative and quantitative evaluations.

% -131
%  58
% -123
% -148
% 54

    % & \multicolumn{1}{c}{User} & \multicolumn{5}{c}{FID} & \multicolumn{5}{c}{Edit distance} \\
    % \cmidrule(lr){2-6}\cmidrule(lr){7-11} Model &
    % & 1-3 & 4-6 & 7-9 & 10-12 & 13+ & 1-3 & 4-6 & 7-9 & 10-12 & 13+ \\
    % \midrule
    % CNN-only & \textcolor{magenta}{0.4} & 3.1 & 8.1 & \textcolor{magenta}{15.8} & 34.7 & \textcolor{cyan}{13.2} & \textcolor{magenta}{26.6} & \textcolor{magenta}{43.6} & \textcolor{magenta}{54.6} & \textcolor{magenta}{90.0} & 0/0 \\
    % GCN & \textcolor{cyan}{0.1} & \textcolor{cyan}{0.8} & \textcolor{cyan}{2.3} & \textcolor{cyan}{3.2} & \textcolor{cyan}{3.7} & \textcolor{magenta}{18.6} & \textcolor{orange}{17.0} & \textcolor{orange}{18.1} & \textcolor{orange}{22.7} & \textcolor{orange}{31.5} & 0/0 \\
    % Ashual \etal ~\cite{ashual2019specifying} & \textcolor{orange}{0.2} & 2.7 & 6.2 & 19.2 & 36.0 & 64.0 & 92.2 & 87.6 & 122.8 & 149.9 & 0/0 \\
    % Johnson \etal ~\cite{johnson2018image} & \textcolor{orange}{0.2} & \textcolor{magenta}{2.6} & \textcolor{magenta}{5.2} & 17.5 & \textcolor{magenta}{29.3} & 69.8 & 86.9 & 80.1 & 117.5 & 123.2 & 0/0\\
    % House-GAN [Ours] & \textcolor{cyan}{0.1} & \textcolor{orange}{1.1} & \textcolor{orange}{2.9} & \textcolor{orange}{3.9} & \textcolor{orange}{10.8} & \textcolor{orange}{13.6} & \textcolor{cyan}{9.4} & \textcolor{cyan}{14.4} & \textcolor{cyan}{11.6} & \textcolor{cyan}{20.1} & 0/0\\

\begin{figure}[tb]
     \centering
     \includegraphics[width=\linewidth]{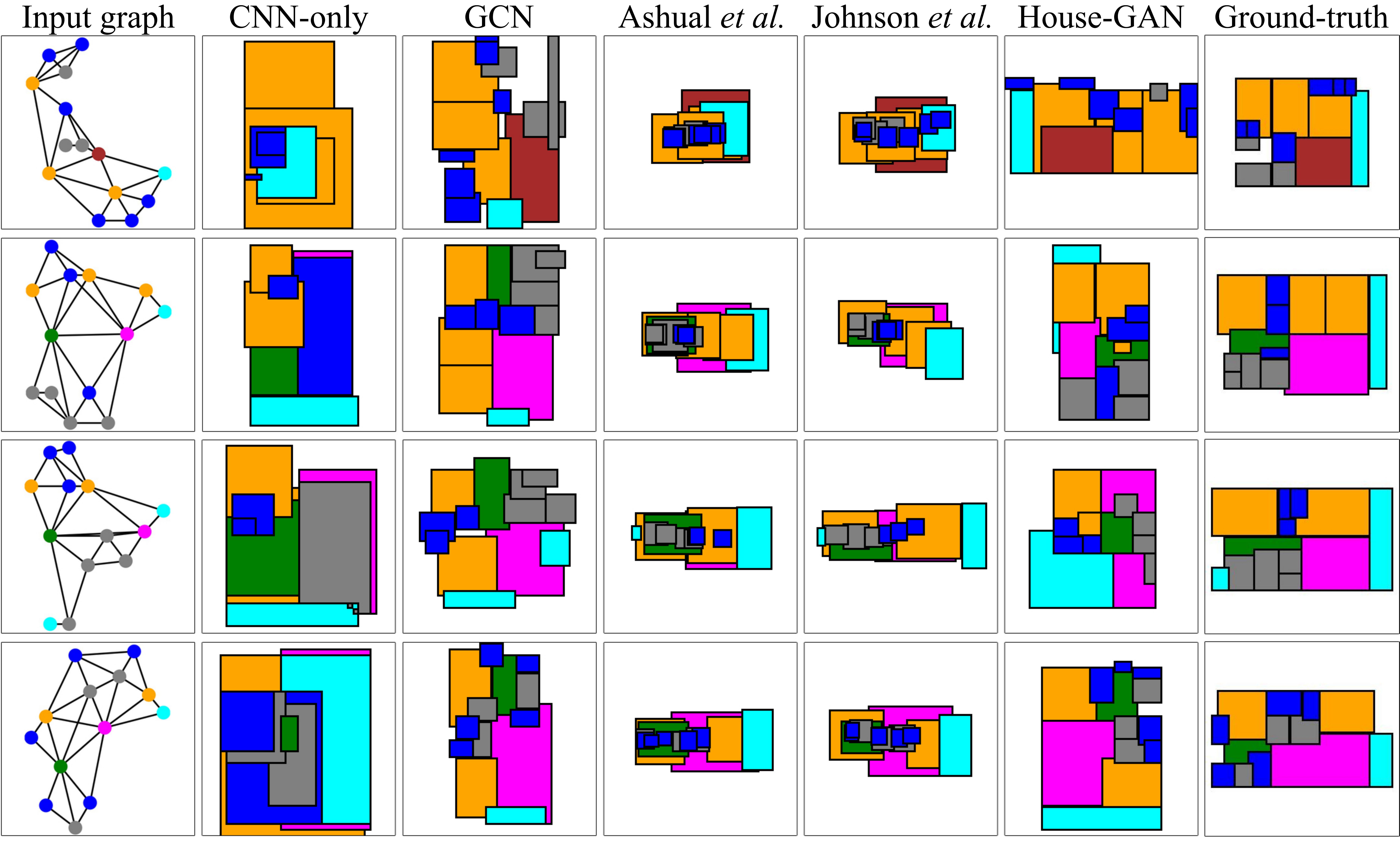}
     \caption{Realism evaluation. We show one layout sample generated by each method from each input graph.
     %     Generated house layout examples.
     Our approach (House-GAN) produces more realistic layouts whose rooms are well aligned and spatially distributed.
     %ve evaluation of realism. Figure shows comparison of generated house layouts from our House-GAN (magenta box) against baselines (CNN-only, GCN) and competing methods for fours different input graphs. All methods were trained and tested on input graphs containing 1-12 and 13+ room nodes, respectively.}
     }
\label{figure:realism}
\end{figure}

%%%%%%%%%%%%%%%%%%%%%%%%%%%%%%%%%%%
% Realism: Qualitative Evaluation %
%%%%%%%%%%%%%%%%%%%%%%%%%%%%%%%%%%%
\mysubsubsection{Realism}
% %%%%%%%%%%%%%%%%%%%%% Main results - User study %%%%%%%%%%%%%%%%%%%%%%%
\begin{figure}[!t]
     \centering
     \includegraphics[width=\linewidth]{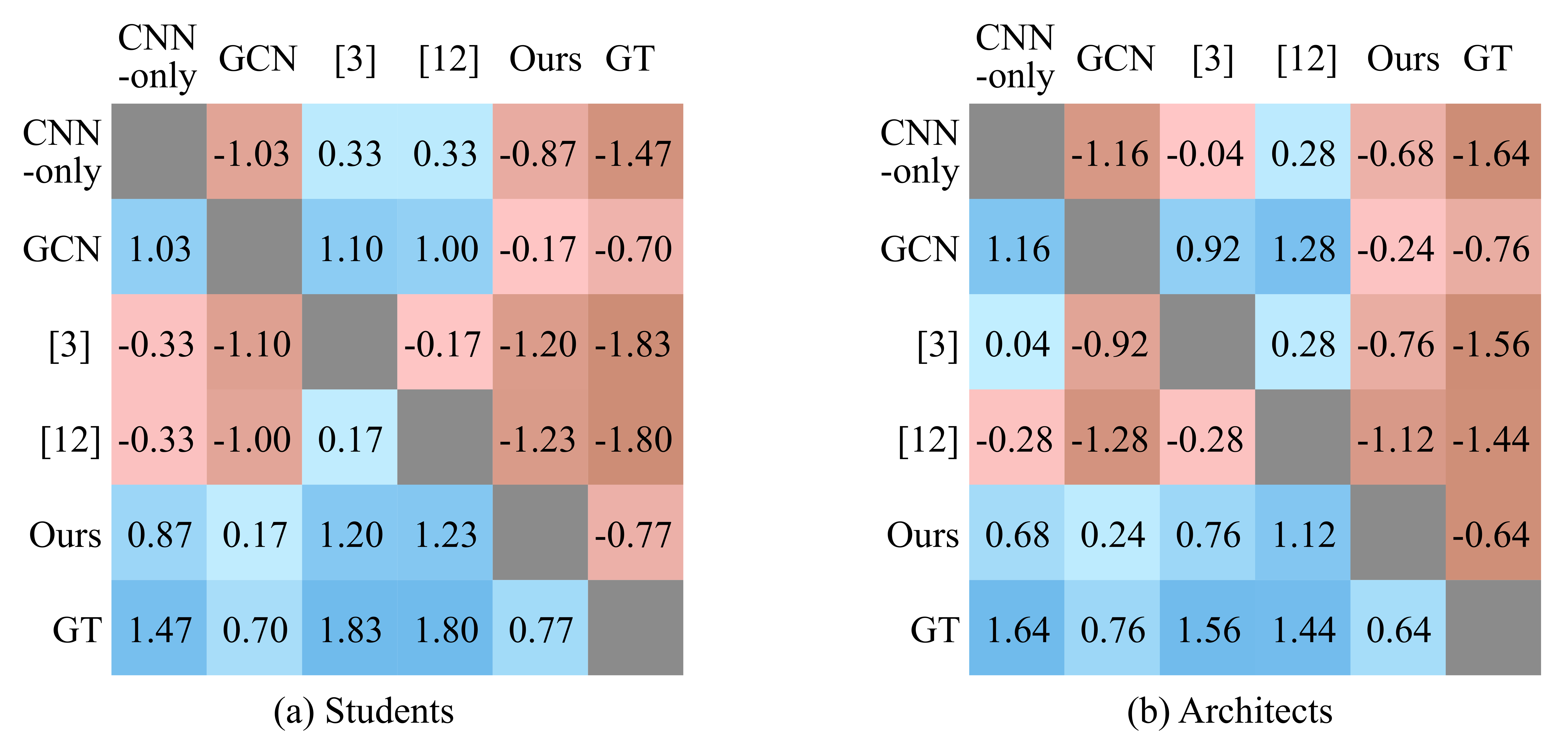}
     \caption{Realism evaluation. The user study score differences for each pair of methods by graduate students (left) and professional architects (right). The tables should be read row-by-row. For example, the bottom row shows the results of GT against the other methods.}
     %The breakdown user study scores by graduate students (left) and professional architects (right) for measuring realism. We compare the baselines, competing methods and ground-truth against each other, reporting scores for each pair of methods.}
\label{fig:user_study}
\end{figure} 
%We present quantitative evaluation of realism on Tables~\ref{tab:main_results} and
We conducted a user study with 12 graduate students and 10 professional architectures. Each subject compared 75 pairs of layouts sampled from the five targeted methods and the ground-truth.
%
%We have six methods to compare (i.e., House-GAN, 4 competing methods, and ground-truth), resulting in 15=$6\choose 2$ pairs for one round of evaluation. We present 5 rounds of questions to each subject and take the average rating as the realism score.
%
%The average user rating by House-GAN is the best as in  Table~\ref{tab:main_results}.
Table~\ref{tab:main_results} shows that House-GAN has the best overall user score.
Figure~\ref{fig:user_study} shows the direct pairwise comparisons. For each pair of methods, we look at the user scores when the 2 methods were compared, compute their average scores, and take the difference.
%In Fig.~\ref{fig:user_study}, for each pair of methods, we look at the user scores when samples from the 2 methods are presented, compute their average scores, and take the difference.
If subjects always choose ``better'' for one method, the difference would be 2.0. Therefore, the score difference of 1.0 (e.g., GCN against \cite{johnson2018image} by students) could mean that the method was rated ``better'' half the time, and ``equally X'' half the time.
%urs against CNN-only is 0.87 by students. This means that students rate ours as ``better'' for a bit less than half the times.
%shows the user rating difference for each pair of methods: We only look at the results when samples from those 2 methods are presented to the subjects
The figure shows that both students and architects rate House-GAN the most realistic except of course the ground-truth.
%House-GAN consistently outperforms all the other methods 
%difference of user ratings for every pair of methods.
%score difference for every pair of methods. Looking at the row for House-GAN, our rating is better than all the other competing methods except for the ground-truth.
%
Figure~\ref{figure:realism} qualitatively supports the same conclusion.
%illustrates that our layouts look the most realistic.
Ashual \etal ~did not produce compelling results, because they rather focus on realistic image generation and need rough locations of objects as input, which are not given in our problem.
%require rough locations of objects as input and rather focus on realistic image generation in their experiments.
%instead of the layout generation.
%\nelson{the etal is missing a space after.}
%
Johnson \etal ~also failed in our experiments. They produce more realistic results if samples from the same group are included in the training set, allowing the method to memorize answers. 
%They employ GCN but a noise vector is added after GCN near the end of the network. 
We believe that their network is not capable of generalizing to unseen cases.

%%%%%%%%%%%%%%%%%%%%%%%%%%%%%%%%%%%%%
% Diversity: Qualitative Evaluation %
%%%%%%%%%%%%%%%%%%%%%%%%%%%%%%%%%%%%%
\begin{figure}[tb]
     \centering
     \includegraphics[width=\linewidth]{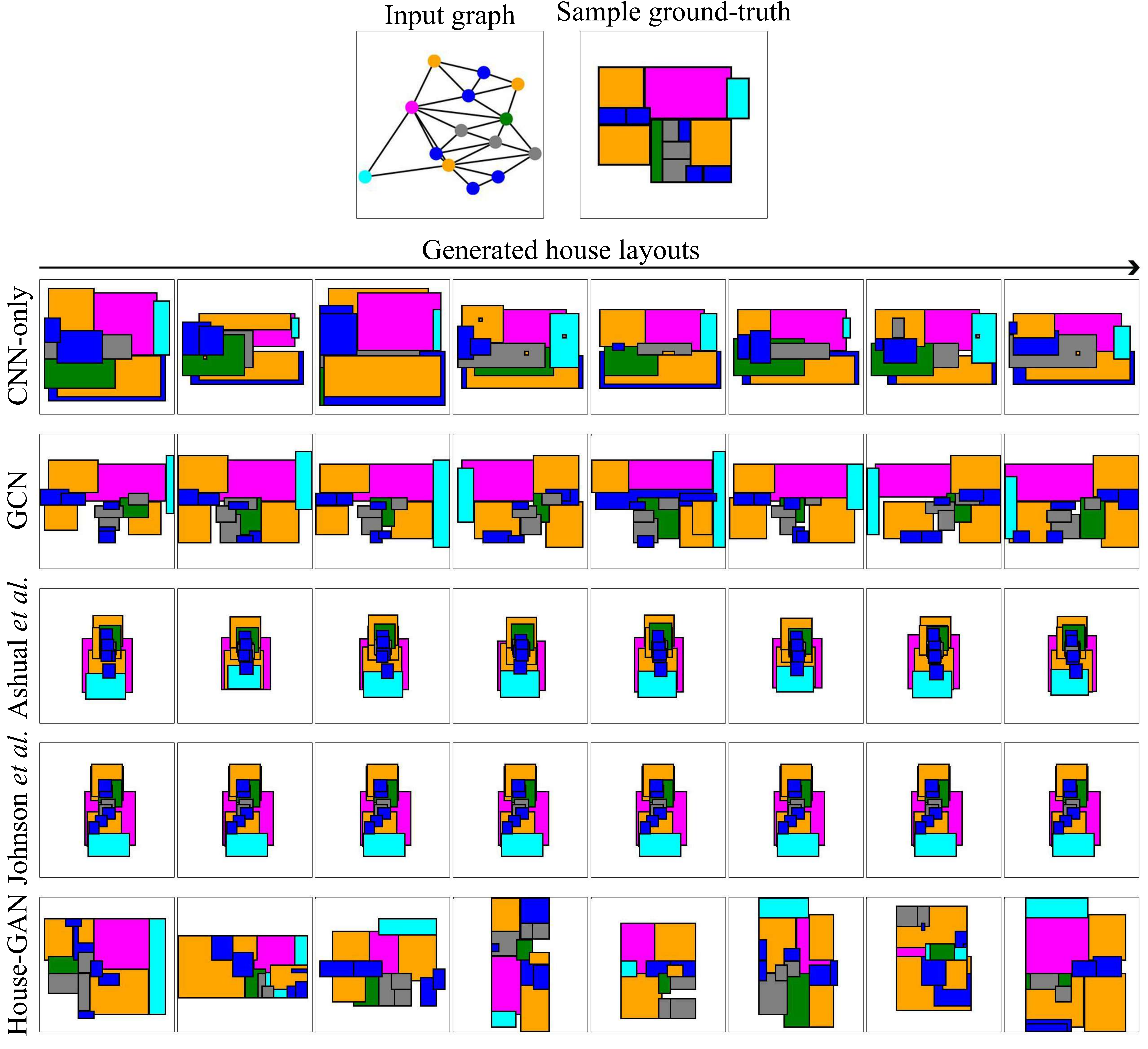}
     \caption{Diversity evaluation: House layout examples generated from the same bubble diagram. House-GAN (ours) shows the most diversity/variations.}
%     Qualitative evaluation of diversity. Figure compares diversity of generated house layouts from our House-GAN against baselines (CNN-only, GCN) and competing methods for the same input graph. All methods were trained and tested on input graphs containing 1-12 and 13+ room nodes, respectively.}
\label{figure:diversity}
\end{figure} 

\mysubsubsection{Diversity}
Diversity is another strength of our approach. For each group, we randomly sample 5,000 bubble diagrams and let each method generate 10 house layout variations. We rasterize the bounding boxes (sorted in a decreasing order of the areas) with the corresponding room colors, and compute the FID score.
%For each of 5,000 randomly sampled bubble diagrams, each method generated 10 layouts
%In Table~\ref{tab:main_results}, we show FID scores for evaluating diversity of results and in
Ashual \etal ~generates variations 
%\nelson{one of the sources according to author}
by changing the input graph into an equivalent form (e.g., apple-right-orange to orange-left-apple). We implemented this strategy by changing the relation from room1-adjacent-room2 to room2-adjacent-room1. However, the method failed to create interesting variations.
%\yasu{fill in what we did to implement this trick}, which did not work at all.
%No constrain
Johnson \etal also fails in the variation metric. Our observation is that they employ GCN but a noise vector is added after GCN near the end of the network. The system is not capable of generating variations.
%and cannot create variations of spatial layouts.
%and is not capable of generating variations in the layout.
%
House-GAN has the best diversity scores except for the smallest group, where there is little diversity and the graph-constraint has little effect.
%succee
%Our idea of encoding the constraint into the relational graph architecture and employing convolutional message passing networks allows our approach to learn to generate interesting spatial variations,
%is really effective in generating 
Figure~\ref{figure:diversity} qualitatively demonstrates the diversity of House-GAN, where the other methods tend to collapse into fewer modes.
%Diversity of 
%which is also qualitatively clear from Fig.~\ref{figure:diversity}.
%, we present our qualitative evaluation. In order to compute the FID score, we rasterize the outputted bounding boxes (sorted in decreasing area size) filled with their corresponding room color. We compute FID score on 10 house layout variations for each of 5,000 bubble-grams from an unseen graph group and use this same set of bubble-grams across all methods. Fig.~\ref{figure:diversity}, shows 8 variations generated by each method for the same given bubble-gram from an unseen graph group and a sample ground-truth house layout. \nelson{add some discussion}

\begin{figure}[!t]
     \centering
     \includegraphics[width=\linewidth]{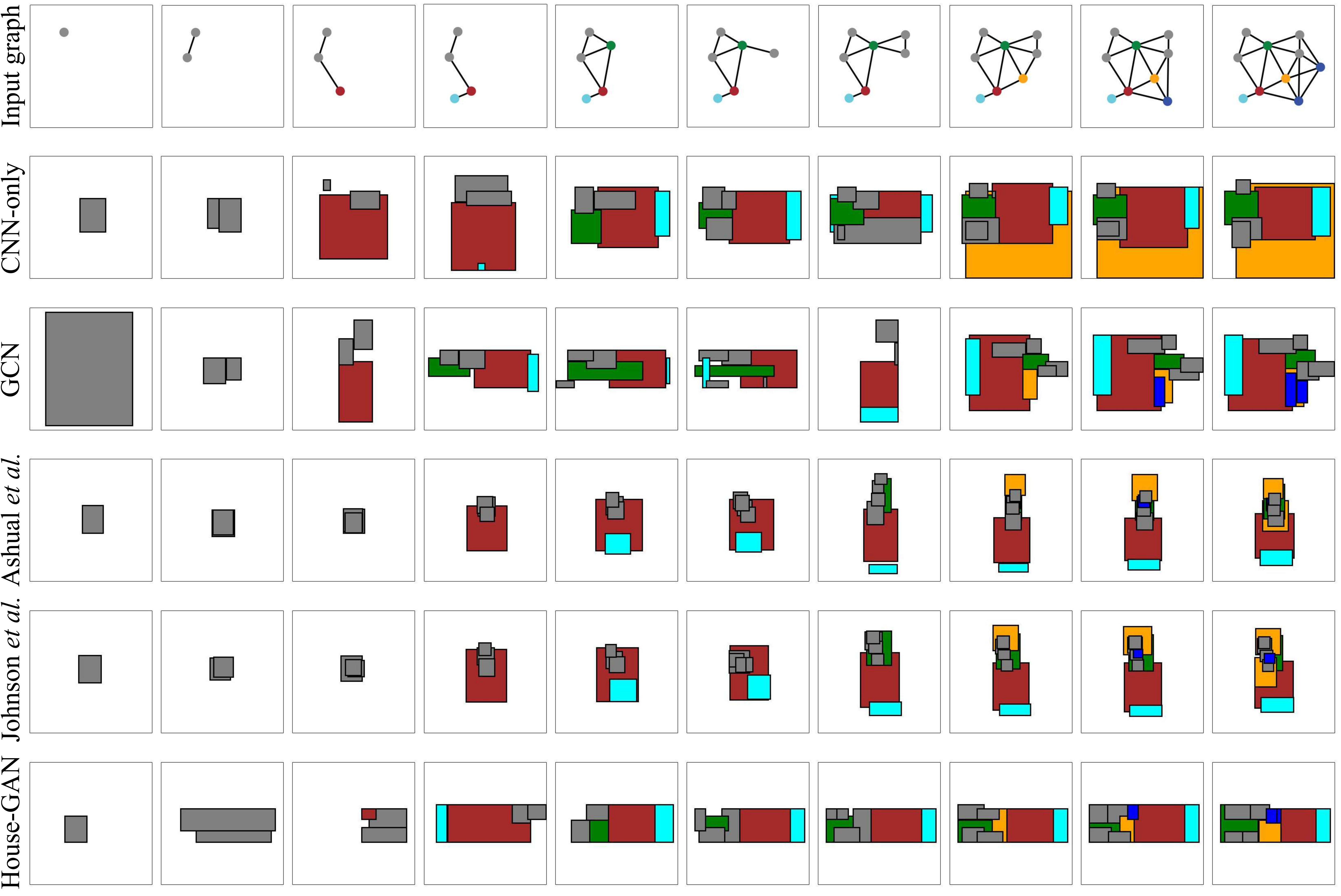}
     \caption{Compatibility evaluation.
%     ity. Figure compares compatibility of generated house layouts from our House-GAN against baselines (CNN-only, GCN) and competing methods.
We fix the noise vectors and sequentially add room nodes one by one with their incident edges. }
%corresponding connections to nodes present in the previous input graph. All methods were trained and tested on input graphs containing (1-9 and 13+) and (10-12) room nodes, respectively.}
\label{figure:compatibility}
\end{figure}

%%%%%%%%%%%%%%%%%%%%% Compatibility: Ablation study %%%%%%%%%%%%%%%%%%%%%%%
\mysubsubsection{Compatibility}
All the methods perform fairly well on the compatibility metric, where many methods collapse to generating a few examples with high compatibility scores. The real challenge is to ensure compatibility while still keeping variations in the output, which House-GAN is the only method to achieve (See Fig.~\ref{figure:compatibility}).
%the real challenge is how to ensure compatibility while still keeping variations in the output
To further validate the effectiveness of our approach, Table ~\ref{tab:ablation} shows the improvements of the compatibility scores as we increase the input constraint information (i.e., room count, room type, and room connectivity).
%from ``no constraints'' to ``room count'', ``room count and types'', and ''entire graph''. \yasu{Change the table/figure? with the new baseline namings here}
%shows our ablation study on compatibility when adding node numbers, types and connectivity as constraints and Fig.~\ref{figure:compatibility} our qualitative evaluation.
%Table ~\ref{tab:ablation} reports the graph edit distance score \cite{abu2015exact} available in NetworkX~\cite{hagberg2008exploring} for evaluating compatibility.
%
%
%The first baseline does not have any graph information, which is similar to the CNN-only architecture.
%Our first baseline follows similar architecture as the CNN-only baseline except for the removal of the input graph and is denoted as ``No constraints". The next baseline, ``With node numbers", utilizes similar architecture as House-GAN except for the removal of the room type and connectivity information (i.e. input graph is fully-connected).
%In ``With node numbers and types" baseline, we add the given node type while maintaining the input graph fully-connected. 
The table demonstrates that House-GAN is able to achieve higher compatibility as we add more graph information.
%exploit input constraint effectively in satisfying the compatibility. 
Figure~\ref{figure:compatibility} demonstrates another experiment, where we fix the noise vectors and incrementally add room nodes one-by-one. It is interesting to see that House-GAN sometimes changes the layout dramatically to satisfy the connectivity constraint (e.g., from the 4th column to the 5th).

%In Fig.~\ref{figure:compatibility}, we sample noise vectors for each method, while keeping them fixed, we add nodes one-by-one for building an input-graph from an unseen graph group in a sequential style. We include connections to existing nodes for a freshly added node as we group the input graph.

% at each room node concatenating it with its noise vector, we call this baseline ``With node numbers and types". Finally, we add node connectivity from the bubble-gram, obtaining the exact same architecture as the proposed House-GAN. Is worth to mention that, since ``No constraints" baseline is not given and does not output the room types, we evaluate it against a given bubble-gram disconsidering them.  

% \nelson{shoun't we show the predicted room segments since that's the direct ouput of the network? Reviewers may ask for that.}
%\nelson{shoun't we show some results on training and testing on the same dataset, since it's what regular GANs do?}    
    
%%%%%%%%%%%%%%%%%%%%%%%%%%%%%%%%%%%%%%%%%
% Compatibility: Qualitative Evaluation %
%%%%%%%%%%%%%%%%%%%%%%%%%%%%%%%%%%%%%%%%%

\begingroup
\renewcommand{\arraystretch}{1.1}
\begin{table*}[!t]
    %scores on methods of increasing input constraint complexity for target graphs of sizes: 1-3, 4-6, 7-9, 10-12 and 13+. 
    \caption{Compatibility evaluation: Starting from the proposed approach that takes the graph as the constraint, we drop its information one by one (i.e., room connectivity, room type, and room count). 
    %The table shows that our framework is capable of achieving higher compatibility as more graph information is added to the input.
    For the top row, where even the room count is not given, it is impossible to form relational networks in House-GAN. Therefore, this baseline is implemented as CNN-only without the room type and connectivity information.
    %    n the room-count is not given, it is impossible to form relational networks of House-GAN. Therefore,     Since House-GAN encodes the graph structure into the relational architecture (i.e., impossible to form our relational networks without knowing the number of rooms), ``no constraints'' baseline is implemented as CNN-only without the room type and connectivity information. 
    The second row (room count only) is implemented as House-GAN while removing the room type information and making the graph fully-connected. Similarly, the third row (room count and type) is implemented as House-GAN while making the graph fully-connected. The last row is House-GAN. We refer to the supplementary document for the full details of the baselines. 
    %exploiting more graph information and improving the compatibility.
    %
    %
    %no input constraint, we add the room-count, the room-type, and the room-connectivity information. House-GAN architecture 
    %The metric (the graph edit distance) with different amount of input information/constraint. Starting from House-GAN that takes an entire graph, we drop room-connectivity, room-type, and room-count information one-by-one.
    %The \textcolor{cyan}{cyan}, \textcolor{orange}{orange}, and \textcolor{magenta}{magenta} colors indicates the first, the second, and the third best results, respectively.
    }
    \label{tab:ablation}
    \centering
    \begin{tabular}{cccccccccccc}
    \toprule
%    &\multicolumn{5}{c}{Edit distance}\\
    %\cmidrule(lr){2-6}
    Count & Type & Conn.
    & 1-3 & 4-6 & 7-9 & 10-12 & 13+ \\
    \midrule
    &&& 28.5 & 28.8 & 19.0 & 26.4 & \textcolor{magenta}{32.2}  \\
    \checkmark&&& \textcolor{magenta}{0.6} & \textcolor{orange}{2.1} & \textcolor{orange}{4.6} & \textcolor{orange}{7.3} & 37.3 \\
    \checkmark&\checkmark&& \textcolor{orange}{0.4} & \textcolor{magenta}{2.2} & \textcolor{magenta}{4.4} & \textcolor{magenta}{7.5} & \textcolor{orange}{21.4}\\ %39.39 \\
    \checkmark&\checkmark&\checkmark& \textcolor{cyan}{0.1} & \textcolor{cyan}{1.1} & \textcolor{cyan}{2.9} & \textcolor{cyan}{3.9} & \textcolor{cyan}{10.8} \\
    \bottomrule
    \end{tabular}
\end{table*}
\endgroup

\begin{figure}[!tb]
\centering
    \includegraphics[width=\linewidth]{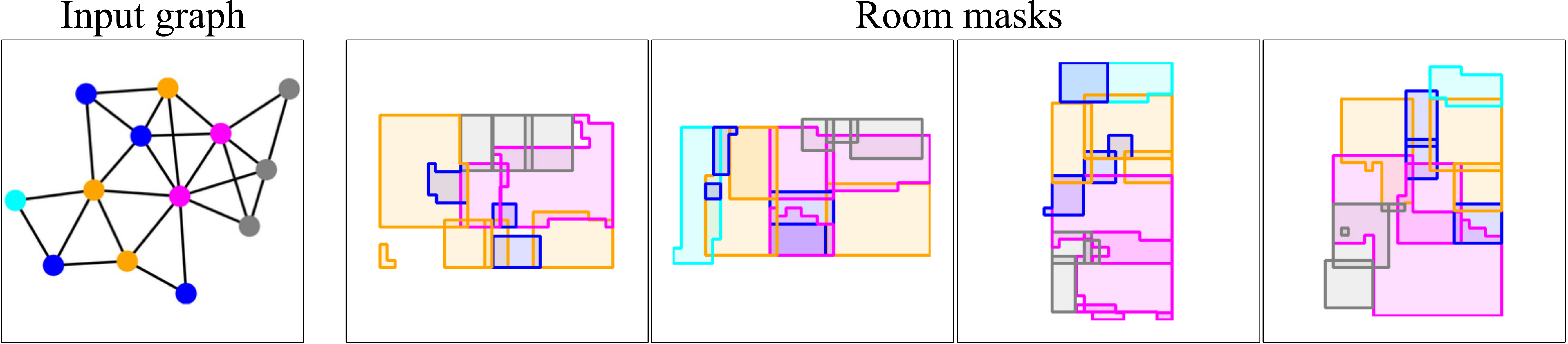}
    \caption{Raw room segmentation outputs before the rectangle fitting.}
    \label{figure:raw_output}
\end{figure}

\mysubsubsection{More results and discussion}
\begin{figure}[!tb]
     \centering
     \includegraphics[width=\linewidth]{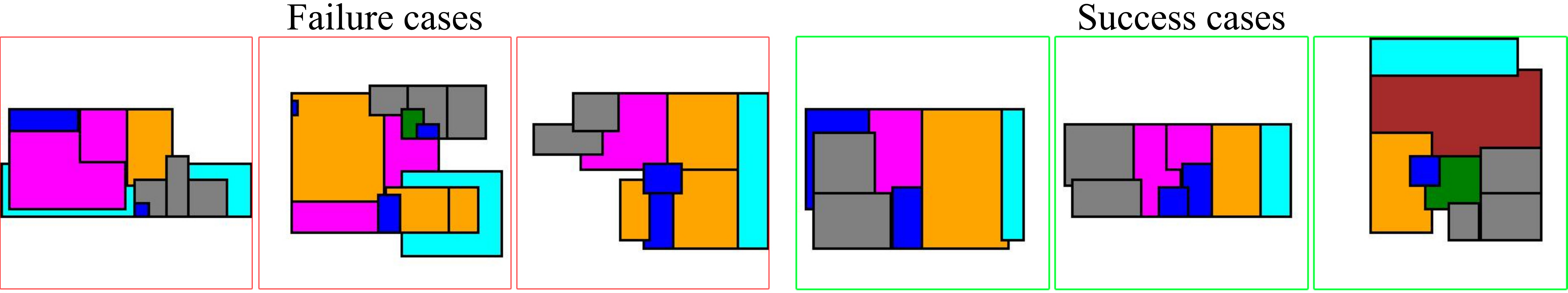}
     \caption{Failure and success examples by House-GAN from the user study. Architects rate the success examples (right) as ``equally good'' and the failure examples (left) as ``worse'' against the ground-truth.}
%     The failure cases (left) consists of samples that scored worse when compared against a ground-truth and the success cases (right) were scored to be equally good to a ground-truth.}
\label{figure:failure_success}
\end{figure} 
Figure~\ref{figure:failure_success} shows interesting failure and success examples that were compared against the ground-truth in our user study.
Professional architects rate the three success examples as ``equally good'' and the three failure cases as ``worse'' against the ground-truth.
For instance, the first failure example looks strange because a balcony is reachabale only through bathrooms, and a closet is inside a kitchen. The second failure example looks strange, because a kitchen is separated into two disconnected spaces. Our major failure modes are 1) improper room size or shapes for a given room type (e.g., a bathroom is too big); 2) misalignment of rooms; and 3) inaccessible rooms (e.g., room entry is blocked by closets). Our future work is to incorporate room size information or door annotations to address these issues. Lastly, Figure~\ref{figure:raw_output} illustrates the raw output of the room segmentation masks before the rectangle fitting. The rooms are often estimated as rectangular shapes, because rooms are represented as axis-aligned rectangles in our dataset, while the original floorplan contains non-rectangular rooms. Another future work is the generation of non-rectangular rooms. We refer to supplementary document for more results.

%failure and success samples from the user study that were scored worse or equally good when compared to a ground-truth house layout, respectively. We noticed three major failure modes in our algorithm as illustrated in Figure~\ref{figure:failure_success}. First, model fails to estimate proper room size and shape for a given room type (e.g. balcony covers entire floorplan). Second common mistake is the presence of some undesired misalignment in the house layout. Lastly, model outputs a disconnected circulation in the layout (e.g. room-to-room access is blocked by closets).         

\section{Conclusion}

This paper proposes a house layout generation problem and
%, whose task is to generate a diverse set of realistic house layouts compatible with a bubble diagram. The paper also presents 
a graph-constrained relational generative adversarial network as an effective solution.
%The key idea is to encode the input constraint into the graph structure of relational generator and discriminator.
%
We define three metrics (realism, diversity, and compatibility) and demonstrate the effectiveness of the proposed system over competing methods and baselines.
%employ relational generator and discriminator, where the input graph constraint is encoded into the their graph structure. We demonstrate that our method makes significant improvement over existing state-of-the-art in three metrics: realism, diversity and compatibility. 
We believe that this paper makes an important step towards computer aided design of house layouts. We will share our code and data.
%to promote further research.

\mysubsubsection{Acknowledgement}
This research is partially supported by NSERC Discovery Grants, NSERC Discovery Grants Accelerator Supplements, and DND/NSERC Discovery Grant Supplement. We would like to thank architects and students for participating in our user study.
% This research is also supported by the Intelligence Advanced Research Projects Activity (IARPA) via Department of Interior / Interior Business Center (DOI/IBC) contract number D17PC00288. The U.S. Government is authorized to reproduce and distribute reprints for Governmental purposes notwithstanding any copyright annotation thereon. The views and conclusions contained herein are those of the authors and should not be interpreted as necessarily representing the official policies or endorsements, either expressed or implied, of IARPA, DOI/IBC, or the U.S. Government.
\clearpage
%\input{paper_planning.tex}
\bibliographystyle{splncs04}
\bibliography{egbib}
\end{document}